\documentclass{article} 
\usepackage[preprint]{colm2026_conference}

\usepackage{microtype}
\usepackage{hyperref}
\usepackage{url}
\usepackage{booktabs}

\usepackage{lineno}

\usepackage{graphicx}
\usepackage{amsmath}
\usepackage{multirow}
\usepackage{color}
\usepackage{hhline}
\usepackage[skins]{tcolorbox}
\usepackage{wrapfig}
\usepackage{xspace}
\usepackage{makecell}
\usepackage{booktabs}
\usepackage{array}
\usepackage{xcolor}
\usepackage{pifont}
\usepackage{colortbl}
\usepackage{adjustbox}
\usepackage{subcaption}
\usepackage{enumitem}

\usepackage[capitalize, nameinlink, noabbrev]{cleveref}

\definecolor{tablegray}{gray}{0.9} 
\newcolumntype{G}{>{\columncolor{tablegray}}c}

\newcommand{\benchmark}{\textsc{MERRIN}\xspace}
\newcommand{\nosearch}{\textit{No Search}\xspace}
\newcommand{\nativesearch}{\textit{Native Search}\xspace}
\newcommand{\agenticsearch}{\textit{Agentic Multimodal Search}\xspace}

\definecolor{darkblue}{rgb}{0, 0, 0.5}
\hypersetup{colorlinks=true, citecolor=darkblue, linkcolor=darkblue, urlcolor=darkblue}

\definecolor{oursgray}{RGB}{228,228,228}
\definecolor{checkgreen}{RGB}{0,148,55}
\definecolor{crossred}{RGB}{204,30,30}
\definecolor{partialyellow}{RGB}{195,130,0}
\definecolor{amethyst}{rgb}{0.6, 0.4, 0.8}

\newcommand{\cmark}{\textcolor{checkgreen}{\ding{51}}}
\newcommand{\xmark}{\textcolor{crossred}{\ding{55}}}

\newcommand{\website}{\raisebox{-1.5pt}{\includegraphics[height=1.05em]{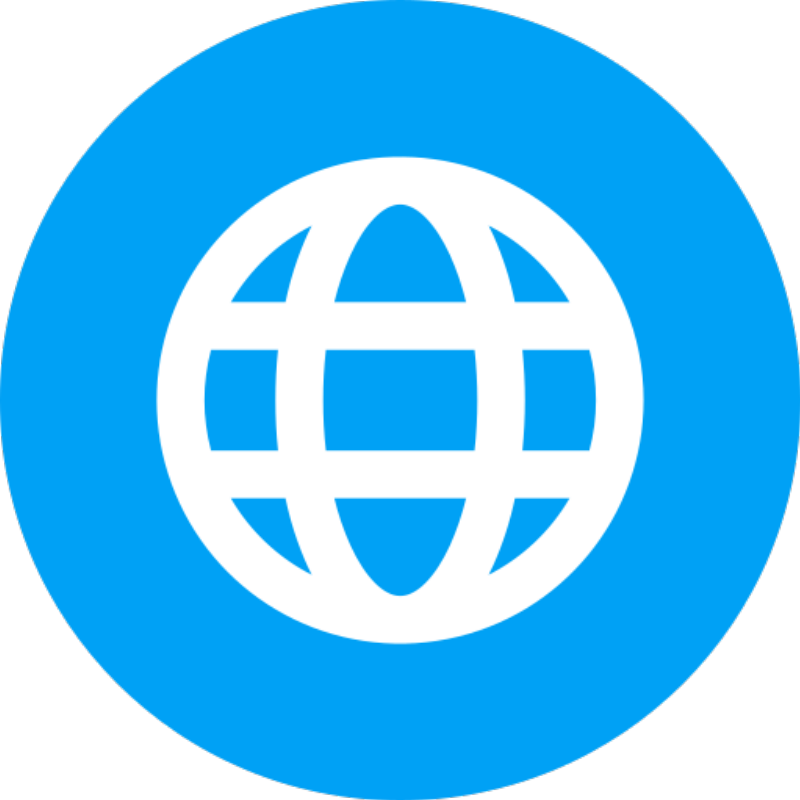}}\xspace}

\title{MERRIN: A Benchmark for Multimodal Evidence Retrieval \\
and Reasoning in Noisy Web Environments}

\author{Han Wang$^{1}$\thanks{Equal contribution. Correspondence to: \texttt{hwang@cs.unc.edu}.} \,\, David Wan$^{1*}$ \,\, Hyunji Lee$^{1*}$ \,\, Thinh Pham$^{2}$ \,\, Mikaela Cankosyan$^{2}$ \,\, \\
\textbf{Weiyuan Chen$^{2}$ \,\, Elias Stengel-Eskin$^{3}$ \,\, Tu Vu$^{2}$ \,\, Mohit Bansal$^{1}$}\\
$^{1}$UNC Chapel Hill \,\, $^{2}$ Virginia Tech \,\, $^{3}$ University of Texas at Austin \\[0.75ex]
\makebox[\textwidth][c]{\website \url{https://merrin-benchmark.github.io}}
\vspace{-12pt}
}

\begin{document}

\ifcolmsubmission
\linenumbers
\fi

\maketitle 

\begin{abstract}
Motivated by the underspecified, multi-hop nature of search queries and the multimodal, heterogeneous, and often conflicting nature of real-world web results, we introduce \benchmark (\underline{M}ultimodal \underline{E}vidence \underline{R}etrieval and \underline{R}easoning \underline{i}n \underline{N}oisy Web Environments), a human-annotated benchmark for evaluating search-augmented agents. 
\benchmark measures AI agents' ability to identify relevant modalities, retrieve multimodal evidence, and perform multi-hop reasoning over noisy web sources. 
It differs from prior work in three important aspects: (1) using natural language queries without explicit modality cues, (2) incorporating underexplored modalities such as video and audio, and (3) requiring the retrieval of complex, often noisy or conflicting multimodal evidence during web search.
We evaluate diverse search agents powered by ten models, including strong closed-source models (e.g., GPT-5.4-mini, Gemini 3/3.1 Flash/Pro) and open-weight models (Qwen3-4B/30B/235B), across three search settings (no search, native search, and agentic search).
Our results show that \benchmark is highly challenging: the average accuracy across all agents is 22.3\%, with the best-performing agent reaching only 40.1\%.
We further observe that while stronger agents like Gemini Deep Research achieve higher performance, gains are modest due to over-exploration; they take more steps and use more tools, but are often distracted by conflicting or partially relevant web content, leading to incorrect answers. 
Our analysis of agent bottlenecks shows that while both search effectiveness and multimodal reasoning remain critical challenges, reasoning is the more pressing limitation.
Compared to humans, these agents consume more resources yet achieve lower accuracy, largely due to inefficient source selection and an overreliance on text modalities.
These findings highlight the need for search agents capable of robust search and reasoning across diverse modalities in noisy web environments, making \benchmark a valuable testbed for evaluating such capabilities. 

\end{abstract}

\section{Introduction}
Knowledge on the web is inherently heterogeneous, spanning text, images, videos, and audio, and is often noisy, incomplete, and conflicting across sources~\citep{xu-etal-2024-knowledge-conflicts,wang2025retrievalaugmented,pham2026sealqa}. Users of search-augmented agents frequently ask questions that require reasoning over multiple modalities, where agents must (1) identify which modalities are necessary and (2) perform multi-hop reasoning over retrieved evidence despite noise and irrelevant information. Evaluating these capabilities requires benchmarks that reflect the complexity of real-world web search, yet prior work has several limitations (\autoref{tab:comparison_searchqa}): many include explicit \textit{modality cues}---direct references to a specific modality such as \emph{``In the following image...''}~\citep{chen-etal-2023-pre-trained,jiang2025mmsearch,li2025mm,zhang2026browsecomp}; the range of modalities is often limited to text and images, excluding video and audio~\citep{jia2025benchmarking,yan2025multimodal,tian2025crosscheckbenchdiagnosingcompositionalfailures}; and the noisy, conflicting nature of real-world web evidence, well-studied in text-only settings~\citep{pham2026sealqa, lee2025corggeneratinganswerscomplex, wang2025retrievalaugmented}, remains underexplored in multimodal settings.

\begin{figure*}[t]
    \centering 
    \includegraphics[width=\linewidth]{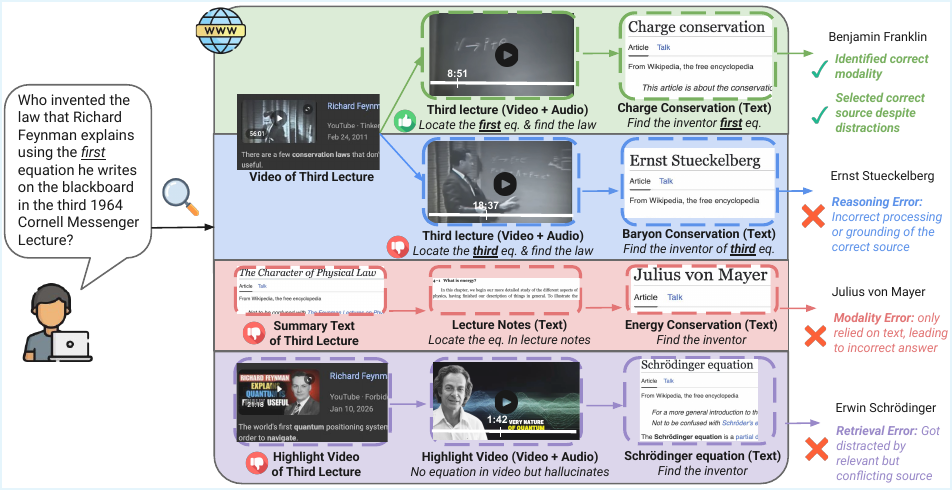}
    \caption{Overview of \benchmark. Given a query, the agent must identify the appropriate modality, retrieve relevant evidence, and perform multi-hop reasoning over noisy, conflicting, and incomplete web sources. The \textcolor{checkgreen}{green} path shows the \textbf{ideal} case: the agent selects the correct modality and source, arriving at the correct answer.
    The remaining paths illustrate three \textbf{failure} modes:
    \textit{Reasoning Error} (\textcolor{blue}{blue})---correct source retrieved but incorrect grounding to the evidence;
    \textit{Modality Error} (\textcolor{red}{red})---agent relies on text when asked about visual information;
    \textit{Retrieval Error} (\textcolor{amethyst}{purple})---correct modality but misleading source selected.} 
    \vspace{-0.5em}
    \label{fig:fig1}
\end{figure*}

To address these gaps, we introduce \benchmark{} (\textbf{M}ultimodal \textbf{E}vidence \textbf{R}etrieval and \textbf{R}easoning \textbf{i}n \textbf{N}oisy Web Environments), a human-annotated benchmark designed to evaluate search-augmented agents under more realistic and challenging conditions.
\benchmark requires agents to identify the necessary modalities and retrieve relevant sources from noisy multimodal evidence on the open web, particularly in scenarios involving multi-hop reasoning across heterogeneous sources and where queries may trigger conflicting, incomplete, or noisy search results.
\cref{fig:fig1} illustrates this challenge.
When an agent correctly identifies both the necessary modalities and the appropriate sources (green path), it can perform accurate reasoning and arrive at the correct answer.
However, even with the correct source, the agent may commit a \textit{Reasoning Error} (blue path): it retrieves the right video but incorrectly grounds the evidence---e.g., locating the \textit{third} equation instead of the \textit{first} in the video---producing an incorrect answer.
A \textit{Modality Error} (red path) occurs when the agent relies on the wrong modality--for instance, using textual evidence when the question requires visual information (e.g., a diagram on a blackboard), leading to incorrect reasoning and answer.
Finally, a \textit{Retrieval Error} (purple path) arises when the agent identifies the right modality but selects a misleading source---e.g., a  summary video rather than the full lecture---and hallucinates evidence that does not exist in the retrieved source.

\begin{table}[t]
\begin{adjustbox}{width=\textwidth,center}
\begin{tabular}{lccccccc}
\toprule
\multirow{2}{*}{\textbf{Benchmark}}
  & \makecell{\textbf{No Explicit} \\ \textbf{Modality Cues}}
  & \makecell{\textbf{Evidence}\\\textbf{Modalities}}
  & \makecell{\textbf{Web Noise}\\\textbf{Reflection}}
  & \makecell{\textbf{Multi}\\\textbf{-hop}} 
  & \makecell{\textbf{Human}\\\textbf{Annotated}} 
  & \makecell{\textbf{Open}\\\textbf{Search}}
\\
\midrule
BrowseComp~\citep{wei2025browsecomp} & - & T& \xmark & \cmark & \cmark & \cmark \\
MM-BrowseComp~\citep{li2025mm} & \xmark & T/I/V & \xmark & \cmark & \cmark & \cmark \\
BrowseComp-VL~\citep{geng2026webwatcher} & \xmark & T/I  & \xmark & \cmark & \xmark & \cmark\\
BrowseComp-V$^3$~\citep{zhang2026browsecomp} & \xmark & T/I & \xmark & \cmark & \cmark & \cmark \\
SealQA~\citep{pham2026sealqa} & - & T & \cmark & \cmark & \cmark & \cmark\\
M3DocVQA~\citep{Cho_2025_ICCV} & - & T/I & \xmark & \cmark & \xmark & \xmark\\
RamDocs~\citep{wang2025retrievalaugmented} & - & T & \cmark & \xmark & \xmark & \xmark \\
MMSearch~\citep{jiang2025mmsearch} & - & T/I & \xmark & \cmark & \cmark  & \xmark\\
MMSearch-Plus~\citep{tao2026mmsearchplus} & \xmark & T/I & \cmark & \cmark & \cmark & \cmark \\
\midrule
\rowcolor{oursgray}
\textbf{\benchmark} & \cmark & T/I/V/A & \cmark & \cmark &\cmark & \cmark \\
\bottomrule
\end{tabular}
\end{adjustbox}
\caption{Comparison of \benchmark with existing benchmarks.
We compare datasets across multiple dimensions: whether queries do not contain explicit modality cues~(\textit{No Explicit Modality Cues}), evidence modalities necessary to answer them~(\textit{Evidence Modalities}), whether questions reflect noisy or conflicting web sources~(\textit{Web Noise Reflection}), whether they require multi-hop reasoning~(\textit{Multi-hop}), whether they are human-annotated~(\textit{Human Annotated}), and whether they support open-web search~(\textit{Open Search}). 
\benchmark uniquely covers all dimensions, supporting multiple evidence modalities across text (\emph{T}), image (\emph{I}), video (\emph{V}), and audio (\emph{A}).
`-' in \textit{No Explicit Modality Cues} indicates settings where modality selection is unnecessary (e.g., controlled or single modality setups).} 
\vspace{-0.5em}
\label{tab:comparison_searchqa}
\end{table}

To evaluate agents on these challenges, we design \benchmark along multiple axes, as shown in \cref{tab:comparison_searchqa}.
The questions are formulated in natural language, without explicit modality cues, requiring agents to autonomously infer which modalities are necessary and retrieve appropriate evidence.
\benchmark{} further expands the scope of modalities to include underexplored sources such as video and audio, alongside more commonly studied modalities like text, images, and tables.
Moreover, inspired by prior observations about the noisy nature of real-world web data in text domain~\citep{wang2025retrievalaugmented, pham2026sealqa, lee2025corggeneratinganswerscomplex}, 
we design our dataset such that each question induces the retrieval of not only relevant documents but also incomplete, conflicting, or misleading distractors. 
For reliable evaluation, we ensure that each question in the dataset has a single unambiguous answer, enabling consistent automatic evaluation of model performance. 

We evaluate search-augmented agents powered by ten different LLMs---seven closed-source (GPT-5.4-nano and -mini~\citep{gpt5.4}, Gemini-3-Flash and -Pro~\citep{gemini3}, Gemini-3.1-Lite and -Pro~\citep{gemini3.1}, and Gemini Deep Research Agent~\citep{geminideepresearch}) and three open-weight (Qwen3-4B, -30B, and -235B~\citep{yang2025qwen3}) models---under three search settings: \nosearch, \nativesearch, and \agenticsearch. Overall, we find that \benchmark is challenging, with an average accuracy of 22.3\% across all runs; even the strongest agent, Gemini-3.1-Pro with \agenticsearch, achieves only 40.1\%. \agenticsearch performs best, averaging 33.7\%, compared to 23.1\% for \nativesearch and 17.3\% for \nosearch (over six models evaluated in all settings).
Notably, increasing the number of search queries or visited pages does not consistently improve accuracy, suggesting that more extensive search does not necessarily translate into better performance.
We further find that more capable agents (e.g., Gemini Deep Research Agent and Gemini Pro \nativesearch) are more prone to over-exploration in noisy web environments, issuing excessive and repeated search queries and tool calls without converging on an answer.

To decouple the sources of error, we analyze whether failures stem from search or reasoning.
Providing annotated gold evidence, thereby removing the need for search, yields only a modest improvement of 7.6\% (40.1\% $\to$ 47.7\%), with performance still remaining relatively low. This suggests that although both search effectiveness and multimodal reasoning remain critical challenges, improving reasoning is the more pressing bottleneck.
In a human evaluation on a 50-example subset, humans achieve 71.4\% accuracy, substantially outperforming the best agentic system (40.1\%), while using fewer resources (nearly 3$\times$ fewer searches) and achieving higher precision in source selection (38.1\% vs.\ 1.8\%). 
However, humans also find the task challenging, with errors often arising from missed or incomplete details in web sources (e.g., incorrect counts or partial answers), highlighting the difficulty of the reasoning component. 
Moreover, humans benefit substantially from additional time (59.2\% $\to$ 71.4\%), whereas agents show diminishing returns (34.0\% $\to$ 40.1\% for \agenticsearch), consistent with the over-exploration pattern: agents issue redundant queries rather than productively deepening their search.
These results highlight the need for stronger search agents that can better assist humans through robust search and reasoning over complex and noisy web environments and effectively integrating diverse modalities.
Overall, \benchmark provides a challenging and realistic testbed for advancing these capabilities.

\section{\benchmark{}}

We present \benchmark{}, a human-annotated benchmark for multimodal evidence retrieval and reasoning, designed to evaluate the ability of search-augmented agents to determine which modalities to retrieve and correctly reason over noisy, conflicting multimodal evidence.
We describe data collection in \cref{dataset: collection} and dataset statistics in \cref{dataset: stats}.

\subsection{Data Collection} \label{dataset: collection}

\paragraph{Question Design.}
\benchmark{} consists of questions governed by three core requirements and additionally classified along two axes.
Every question must satisfy: (1)~\textit{no modality cues}---questions are phrased in natural language without explicit modality references (e.g., \emph{``shown in the image''}), resembling realistic user queries; (2)~\textit{non-text evidence required}---each question is manually verified to require non-text evidence, with no text-only shortcut available; and (3)~\textit{unique, verifiable answers}---each question has exactly one correct, short, and unambiguous answer.
Each question is further classified along two axes. \textit{Reasoning type} (one or both): multi-hop reasoning (combining information across sources or modalities) or multimodal conflict resolution (reconciling inconsistent evidence across modalities triggered empirically in real search engines; no synthetic conflicts). \textit{Multimodal role} (one or both): non-text evidence may serve as the answer source (the answer can only be extracted from a non-text source) or as a reasoning component (non-text evidence provides an intermediate fact necessary to derive the final answer).
Most questions and evidence are generated from scratch, while some are adapted from SealQA~\citep{pham2026sealqa} and ChartMuseum~\citep{tang2025chartmuseum}, using their question--answer pairs as one hop and augmenting with additional evidence to construct \textit{new} multi-hop questions.
For each question, annotators record the ground-truth answer with reasoning steps, source URLs, source types, multimodal role, reasoning type, and question origin. Full annotation details and guidelines are in \cref{app: data_collection_details}.
\vspace{-0.5em}

\paragraph{Quality Control.}
We employ a multi-round human review process. Each question is reviewed by a second annotator for answer correctness, question clarity, question difficulty, and non-text modality requirements. In the first round, approximately 39.5\% of candidates were rejected; of those, 45.3\% were successfully revised and accepted in the second round.
To verify \textbf{non-text modality requirements}, we decompose each question into sub-questions and attempt to answer each via text-only Google Search. We then perform an adversarial search pass, querying each sub-question together with the known answer to check for text-only shortcuts. A question passes only if at least one sub-question resists both search passes. Further details are in \cref{app: quality_control_details}.
\vspace{-0.5em}

\paragraph{Human Annotators.}
Questions were constructed and reviewed by five graduate-level and one undergraduate annotators with NLP backgrounds. Six annotators constructed questions and four conducted quality control; no annotators reviewed their own questions. Annotators were provided with detailed guidelines~(\cref{app:human_annotation}) and diverse exemplars.

\begin{figure}
    \centering
    \begin{subfigure}[b]{0.32\textwidth}
    \includegraphics[width=\textwidth]{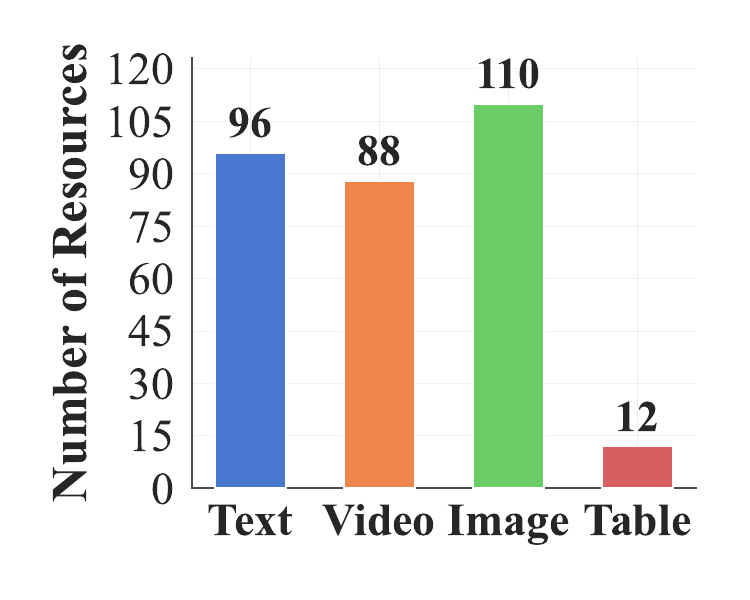}
    \caption{Source Types.}
    \end{subfigure}
    \begin{subfigure}[b]{0.32\textwidth}
    \includegraphics[width=\textwidth]{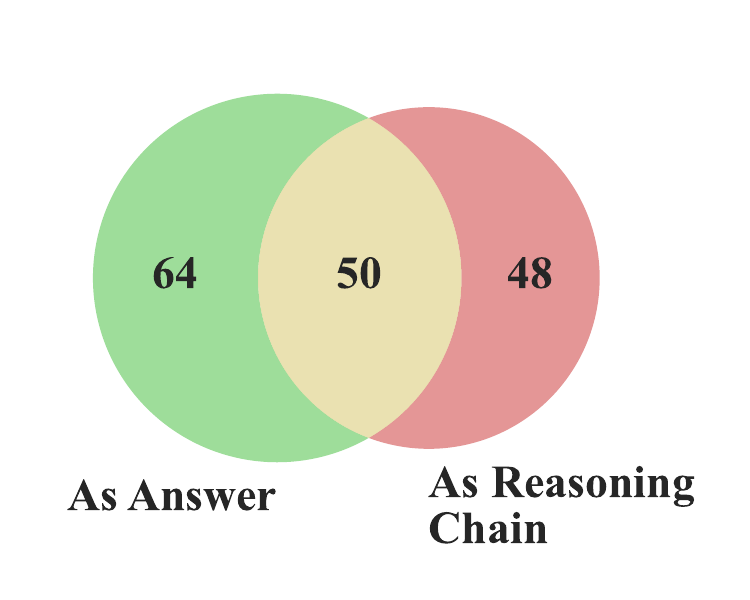}
    \caption{Multimodal Role.}
    \end{subfigure}
    \begin{subfigure}[b]{0.32\textwidth}
    \includegraphics[width=\textwidth]{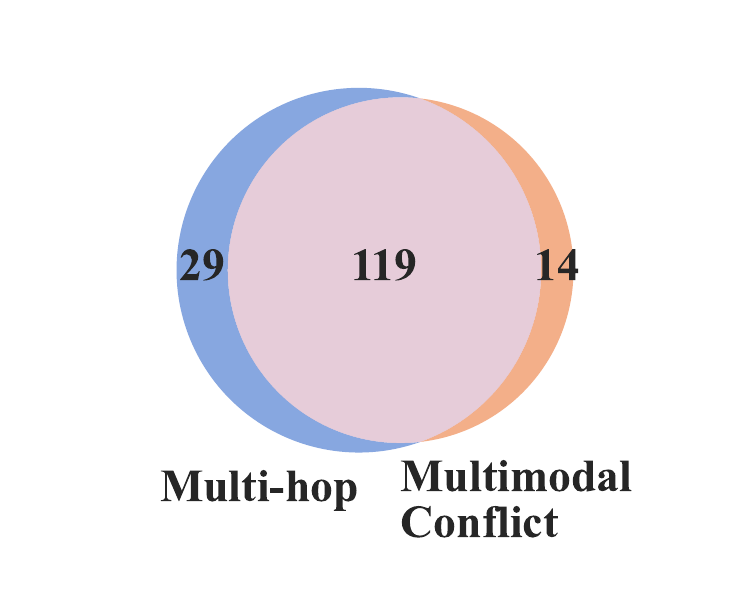}
    \caption{Reasoning Type.}
    \end{subfigure}
    \caption{\benchmark composition. (a) Gold source resources by modality. (b) Questions by the role of visual content. (c) Questions by reasoning type.}
    \vspace{-1em}
    \label{fig:dataset_composition}
\end{figure}

\vspace{-0.5em}
\subsection{Data Statistics} \label{dataset: stats}

\benchmark{} comprises 162 questions\footnote{We focus on a high-quality, expert-vetted diagnostic benchmark, comparable in scale to SealQA's SEAL-0 (111 questions)~\citep{pham2026sealqa} and GPQA-Diamond (198 questions)~\citep{rein2023gpqagraduatelevelgoogleproofqa}.} (120 from scratch, 37 from SealQA, 5 from ChartMuseum).
Four source types (text, image, video, and table) are represented, with text and image most prevalent (\cref{fig:dataset_composition}a). Non-text evidence serves as both answer sources and reasoning components in comparable proportions (\cref{fig:dataset_composition}b), and 73.5\% of questions require both multi-hop reasoning and multimodal conflict resolution (\cref{fig:dataset_composition}c); see \cref{tab:dataset-stats} in the Appendix for full statistics.

\section{Experiments}

\subsection{Setup}

\paragraph{Search-Augmented Agents.} 
We evaluate search-augmented agents powered by ten different models, including closed-source models (GPT-5.4-Nano and -Mini~\citep{gpt5.4}, Gemini-3-Flash and -Pro~\citep{gemini3}, Gemini-3.1-Flash-Lite and -Pro~\citep{gemini3.1}, and Gemini-Deep-Research Agent~\citep{geminideepresearch}), and open-weight models (Qwen 3~\citep{yang2025qwen3} at three scales: 4B, 30B, and 235B), under three search settings: \nosearch (no search tool), \nativesearch (enable each model's built-in search tools), and \agenticsearch (a multimodal search agent framework built using smolagents~\citep{smolagents}).
\nativesearch often does \textit{not} support video and audio processing when accessed via built-in search tools~(\cref{tab:modality_support} in Appendix).
\agenticsearch equips models with various tools to operate across \textit{all} modalities. 
Full details on model configurations, search tools, and the \agenticsearch framework are in \cref{app: exp: setup}.

\paragraph{Metrics.}
We measure \textit{accuracy}, i.e., whether the predicted answer matches the ground truth, using an LLM-as-judge following BrowseComp~\citep{wei2025browsecomp}.\footnote{Manual inspection on 50 instances finds all judgments correct. Due to \benchmark's unambiguous design, most cases reduce to an exact match with text normalization.}
 
\begin{table}[t]
\centering
\resizebox{\textwidth}{!}{
\begin{tabular}{l|G|Gcc|Gcc}
\toprule
& \multicolumn{1}{c|}{\nosearch} & \multicolumn{3}{c|}{\nativesearch} & \multicolumn{3}{c}{\agenticsearch}\\
 \midrule
\textbf{Model} & \textbf{Acc} & \textbf{Acc} & \textbf{\# Search Qs} & \textbf{\# Pages} & \textbf{Acc} & \textbf{\# Search Qs} & \textbf{\# Pages}\\
\midrule
Qwen3-4B & $10.3_{\pm0.4}$ & - &  - & - & $10.5_{\pm1.6}$ & $1.7_{\pm0.0}$ & $0.2_{\pm0.1}$ \\
Qwen3-30B & $8.0_{\pm0.6}$ & - & - & - & $16.1_{\pm0.6}$ & $2.0_{\pm0.1}$ & $0.5_{\pm0.0}$ \\
Qwen3-235B & $12.1_{\pm0.9}$ & - & - & - & $23.3_{\pm1.3}$ & $3.0_{\pm0.2}$ & $1.0_{\pm0.1}$ \\
\midrule
GPT-5.4-nano & $9.9_{\pm2.7}$ & $12.6_{\pm1.3}$ & $37.7_{\pm2.4}$ & $5.9_{\pm0.4}$ & $31.9_{\pm3.0}$ & $11.6_{\pm0.4}$ & $7.5_{\pm0.4}$ \\
GPT-5.4-mini & $14.0_{\pm0.4}$ & $15.6_{\pm0.4}$ & $38.6_{\pm3.7}$ & $5.3_{\pm0.2}$ & $31.1_{\pm3.1}$ & $9.2_{\pm0.3}$ & $3.4_{\pm0.2}$ \\
Gemini 3 Flash & $19.1_{\pm3.2}$ & $31.7_{\pm3.8}$ & $44.1_{\pm0.6}$ & $0.1_{\pm0.0}$ & $32.9_{\pm0.9}$ & $14.8_{\pm0.5}$ & $1.4_{\pm0.0}$ \\
Gemini 3 Pro & $23.5_{\pm1.1}$ & $28.8_{\pm1.4}$ & $34.9_{\pm2.0}$ & $0.1_{\pm0.0}$ & $39.9_{\pm1.6}$ & $8.4_{\pm0.3}$ & $3.0_{\pm0.1}$ \\
Gemini 3.1 Lite & $12.8_{\pm2.3}$ & $20.6_{\pm2.2}$ & $19.2_{\pm1.1}$ & $0.0_{\pm0.0}$ & $26.3_{\pm1.9}$ & $8.3_{\pm0.1}$ & $0.8_{\pm0.1}$ \\
Gemini 3.1 Pro & $\textbf{24.7}_{\pm1.6}$ & $29.0_{\pm1.1}$ & $35.8_{\pm0.9}$ & $0.1_{\pm0.0}$ & $\textbf{40.1}_{\pm2.8}$ & $8.6_{\pm0.3}$ & $2.9_{\pm0.0}$ \\
\midrule
\multicolumn{1}{c|}{Gemini Research} & - & $\textbf{33.3}_{\pm2.2}$ & - & - & - & - & - \\
\bottomrule
\end{tabular}
}
\caption{
Performance of search agents powered by different models on \benchmark.
\textbf{Acc} denotes average accuracy over three runs with standard deviation, \textbf{\# Search Qs} is the average number of search queries issued per question, and \textbf{\# Pages} is the average number of webpages explicitly visited and read per question. \nosearch has no search module; thus, both \# Search Qs and \# Pages are 0.
Gemini 3.1 Lite refers to Gemini 3.1 Flash Lite, and Gemini Research refers to the Gemini Deep Research Agent.
For Qwen models, \nativesearch is not applicable since they do not have an internal search agent. 
Gemini Research only supports using its built-in search system~(\nativesearch) and detailed outputs are unavailable, so \# of Search Qs and \# Pages are omitted.
}
\label{tab:main_results}
\end{table}

\subsection{Results}
\cref{tab:main_results} presents the performance of search-augmented agents powered by ten models on \benchmark under three search settings, with results averaged over three runs.

\paragraph{Overall Performance.}
Overall, the task is challenging for all agents, with an average accuracy of 22.3\% across all runs.
When averaging over the six models evaluated in all three settings, models achieve only 17.3\% accuracy in \nosearch, indicating that \benchmark cannot be solved using parametric knowledge alone; even the strongest model, Gemini-3.1-Pro, reaches only 24.7\%.
Performance improves to 23.1\% with \nativesearch, where agents rely on built-in search pipelines, with the Gemini Deep Research Agent achieving the highest accuracy at 33.3\%.
Performance further increases to 33.7\% with \agenticsearch, which enables access to \textit{all} multimodal evidence, unlike \nativesearch, which does not support video during search~(Appendix~\ref{app: exp: setup}), highlighting the importance of flexible evidence integration across various modalities.
The best overall result is achieved by Gemini-3.1-Pro with \agenticsearch (40.1\%), suggesting that both strong models and robust search frameworks are critical.
Comparing model families, GPT-based agents perform substantially worse than Gemini agents under \nativesearch, with an absolute gap of 13.4\%. However, this gap narrows to 3.3\% under \agenticsearch, suggesting that more flexible search capabilities help close the gap between model families.

\paragraph{Performance of Search Agents with Closed vs. Open-Weight Models.} We observe that \benchmark is particularly challenging for agents powered by open-weight models, the Qwen series, which achieve an average accuracy of 16.6 even when using \agenticsearch, where external search is enabled.
Although both agent types use the same tools and therefore retrieve similar evidence and interpretations, while access to external evidence improves performance for closed-source agents, the gains for open-weight agents are limited, with an average improvement of 16.4 from \nosearch to \agenticsearch for closed agents compared to only 6.5 for open-weight agents.
We attribute the limited gains for open-weight models to three main factors: (1) failure to effectively process long, multi-step search results; (2) greater susceptibility to distraction from irrelevant evidence, leading to premature termination even when the generated answer is incorrect; and (3) weaker reasoning ability, which leads to incorrect intermediate reasoning that propagates to incorrect final answers.

\paragraph{Average Search Queries and Pages Visited.}

For each agent and search setting, we analyze the average number of search queries issued (\textit{\# Search Qs}) and the average number of pages visited (\textit{\# Pages}).
We observe that these metrics are not strongly correlated with accuracy~(\textit{Acc}). 
In particular, a higher number of search queries or visited pages does not necessarily lead to better performance.
Similar trends are observed across both \nativesearch and \agenticsearch. The highest accuracy is achieved by Gemini-3.1-Pro agent, while the largest number of search queries is observed for Gemini-3 Flash agent, and the highest number of pages visited is observed for GPT-5.4-nano agent.

\subsection{Analysis of Failure Modes}\label{sec:failure_modes}
We provide a detailed quantitative and qualitative analysis of the best performing agent in \cref{tab:main_results}, Gemini-3.1-Pro.
\paragraph{Bias Toward Text Modality.}
We observe that agents exhibit a strong bias toward retrieving textual evidence, often failing to identify the most appropriate modality for a given query. 
Specifically, 87.7\% of retrieved evidence is text, compared to only 6.8\% from images and 5.5\% from video and audio combined. 
In contrast, the dataset distribution is more balanced, with 31.4\% text, 35.9\% image, and 28.8\% for video and audio.
This discrepancy indicates that, although text is the dominant modality in the dataset, it is significantly preferred by search agents in retrieved evidence, often leading to incorrect answers.

\paragraph{Error Propagation in Multi-Step Retrieval.}
To analyze multi-step retrieval, we construct 50 human-annotated examples in which each question requires a two-step reasoning chain. For each example, annotators provide sub-questions and intermediate answers for each step. 
We then analyze where agents fail by evaluating whether they correctly produce these intermediate answers. 
Among incorrect predictions, the first step is more often the point of failure (57.7\%) than the second step (42.3\%), indicating that the initial evidence identification is a frequent source of error and that early errors often propagate, leading to incorrect final answers. 
Second step failures are mostly associated with \textit{as answer} multimodal instances (63.6\%), compared to \textit{reasoning chain} (18.2\%) and \textit{both} (18.2\%), indicating the difficulty of understanding and integrating multimodal information to produce the final answer. 

\paragraph{Analysis Across Dataset Axes.}
We find that performance is similar when non-text modalities are required in the \textit{reasoning chain} (45.8\%) and \textit{as answer} (45.3\%), but drops substantially to 28.0\% when both are required. We further examine performance across different question types. 
Performance is higher on multi-hop questions (55.2\%) and multimodal conflict questions (57.1\%), but decreases to 34.5\% when both challenges are present. This mirrors the trend above, indicating that the combination significantly increases task difficulty.

\paragraph{Over-Exploration in Noisy Web Environments.}

We observe that more capable agents (e.g., Gemini Deep Research Agent and Gemini Pro \nativesearch) frequently over-explore when confronted with noisy web evidence, spending excessive time or issuing excessive tool calls without converging on an answer.
Gemini Deep Research times out on an average of 33.1\% of questions, continuing to iteratively search and read for up to 15 minutes without producing a final answer---getting lost in the noise of conflicting or tangentially relevant web content.
A similar pattern emerges for Gemini Pro models under \nativesearch{}: an average of 12.7\% of questions trigger \textsc{Too\_Many\_Tool\_Calls}, where the model exceeds the API's internal limit on search invocations, resulting in empty responses. In contrast, Flash and Lite variants are far less affected (3.1\% and 0.4\%, respectively), as they issue fewer queries and converge more quickly.
This suggests a counterintuitive trade-off: more capable agents are more prone to over-exploration, issuing more search queries in an attempt to gather comprehensive evidence, but ultimately failing to answer within platform constraints.

\section{Additional Analysis}

\begin{figure}[t]
\centering
\begin{minipage}{0.49\linewidth}
    \centering
        \includegraphics[width=\linewidth]{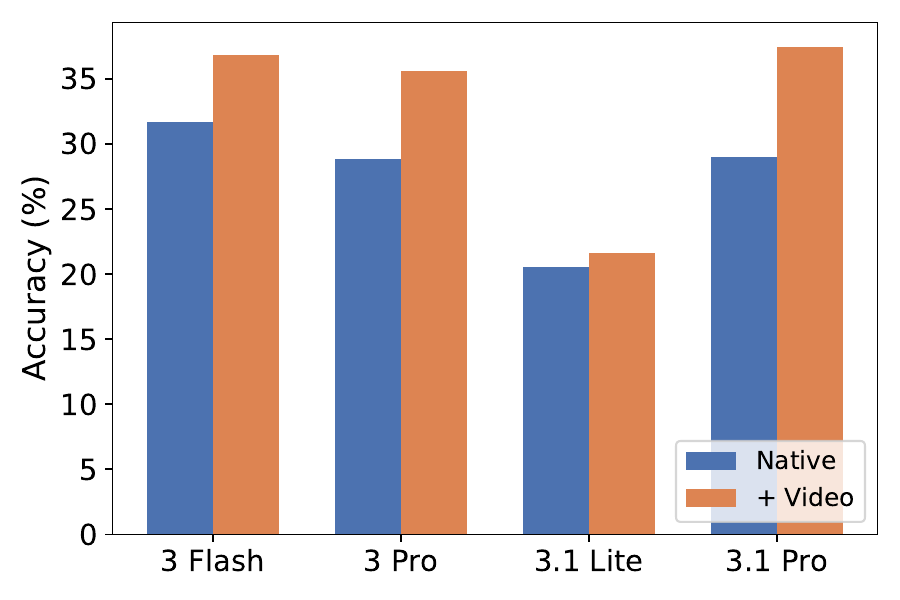}
        \caption{Performance of \nativesearch (blue), when adding video tool (orange).}
        \label{fig:video}
\end{minipage}
\hfill
\begin{minipage}{0.49\linewidth}
    \centering
        \includegraphics[width=\linewidth]{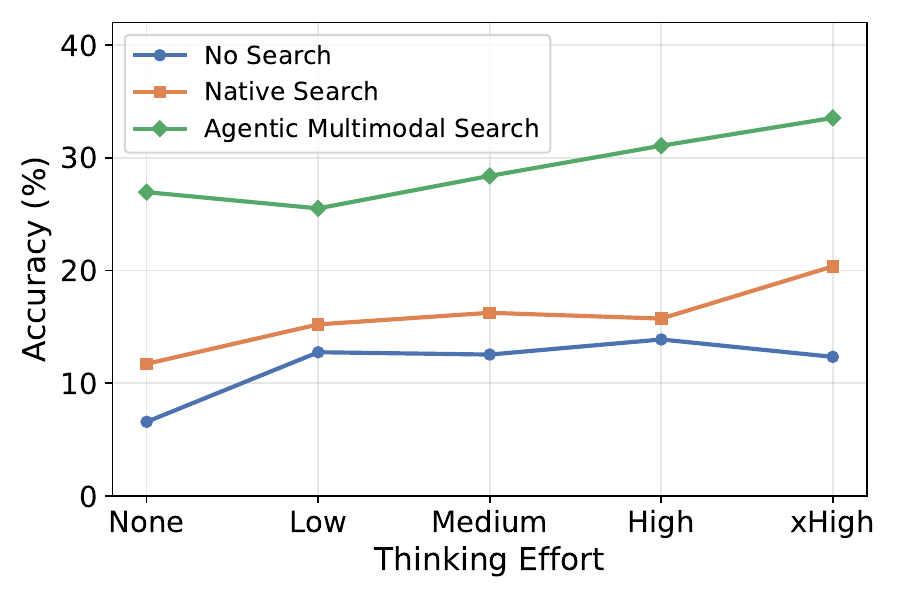}
        \caption{Accuracy of GPT-5.4-mini across different thinking efforts.}
        \label{fig:thinking}
\end{minipage}
\end{figure}

\subsection{Impact of Adding Video Processing Tool}
As \nativesearch is limited to text and image modalities and cannot process video or audio during search~(\cref{tab:modality_support} in Appendix), we investigate the effect of augmenting it with a video processing tool.
As shown in \cref{fig:video}, we observe that when adding a video processing tool, the performance consistently increases with an average of 5.7\% absolute improvement over the four Gemini agents (Gemini 3 Flash/Pro, Gemini 3.1 Lite/Pro), highlighting the importance of enabling access to a broader range of modalities for effective multimodal reasoning. More analysis in \cref{app: ablation: video_process}.

\subsection{Impact of Thinking Effort}

To analyze how varying levels of thinking effort affect performance on \benchmark, we conduct experiments across three search frameworks using GPT-5.4-mini.\footnote{The analysis is conducted with GPT-5.4-mini, as Gemini series does not support disabling thinking.} 
We observe that performance generally improves as thinking effort increases, with the largest gains observed in \agenticsearch, which shows an absolute improvement of 8.6\% when comparing no thinking to the highest level of thinking effort. \nativesearch follows with a 6.8\% improvement, while \nosearch shows a smaller gain of 3.1\%.

\begin{figure}[t]
\centering
\begin{minipage}{0.49\linewidth}
    \centering
    \includegraphics[width=\linewidth]{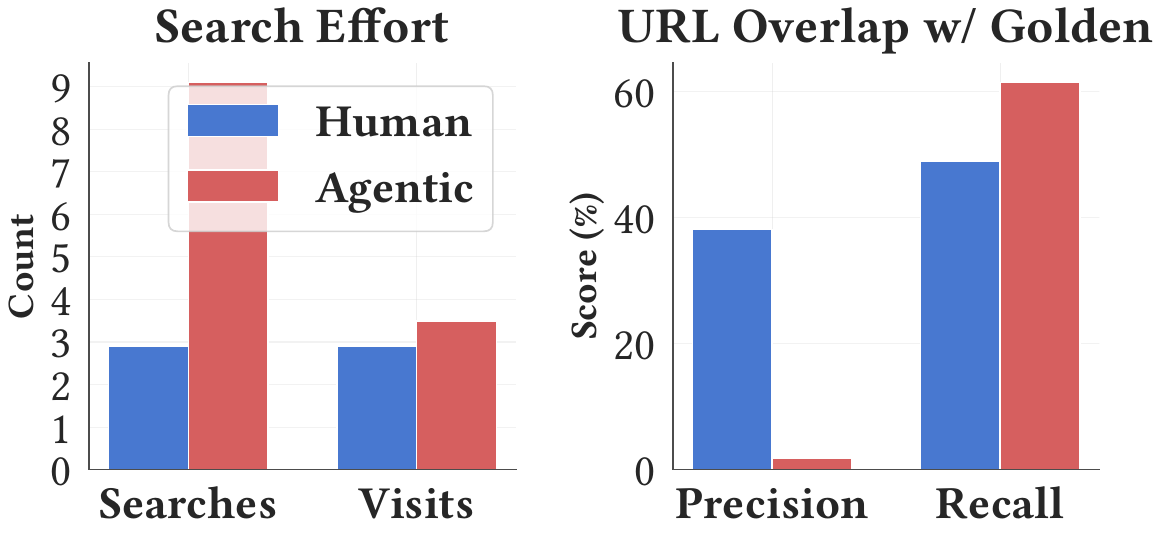}
    \caption{Search effort and URL overlap comparison between humans and \agenticsearch (Gemini-3.1-Pro).}
    \label{fig:human_agentic}
\end{minipage}
\hfill
\begin{minipage}{0.49\linewidth}
    \centering
    \includegraphics[width=\linewidth]{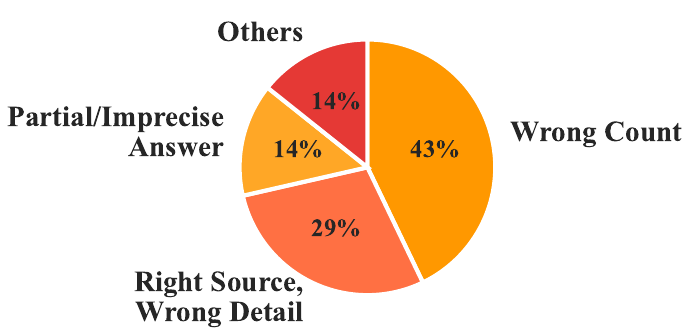}
    \caption{Distribution of human error types, categorized by failure mode.}
    \label{fig:human_error}
\end{minipage}
\end{figure}

\begin{table*}[t]
\centering
\resizebox{\textwidth}{!}{
    \begin{tabular}{l ccc c ccc c ccc c ccc}
    \toprule
    & \textbf{Acc.} & \textbf{Acc@5min} & \textbf{Time}
    & & \multicolumn{3}{c}{\textbf{Search Effort}}
    & & \multicolumn{3}{c}{\textbf{Modality (\%)}}
    & & \multicolumn{3}{c}{\textbf{URL Overlap w/ Golden}} \\
    \cmidrule(lr){2-2} \cmidrule(lr){3-3} \cmidrule(lr){4-4} \cmidrule(lr){6-8} \cmidrule(lr){10-12} \cmidrule(lr){14-16}
    \textbf{System}
    & (\%) & (\%) & (min)
    & & Searches & Visits & URLs
    & & Text & Video & Image
    & & Prec. & Rec. & F1 \\
    \midrule
    Human   & 71.4 & 59.2 & 4.1 && 2.9 & 2.9 & --- && 53.2 & 28.2 & 18.5 && 38.1 & 48.9 & 42.8 \\
    Native  & 30.9 & 29.6 & 2.3 && 9.8 & 0.1 & 34.9 && 96.2 & 0.0 & 3.8 && --- & --- & --- \\
    Agentic & 40.1 & 34.0 & 4.0 && 9.1 & 3.5 & 63.6 && 87.0 & 4.4 & 8.5 && 1.8 & 61.4 & 3.6 \\
    \bottomrule
    \end{tabular}
}
\caption{Performance across human annotators, \nativesearch~(Native), and \agenticsearch~(Agentic) with Gemini 3.1 Pro. \textbf{Acc@5min}: accuracy under a 5-minute budget, where any question taking more than 5 minutes is counted as incorrect. \textbf{Time}: average completion time in minutes. \textbf{Search Effort}: average number of search queries issued (Searches), webpages visited (Visits), and unique URLs encountered (URLs) per question. \textbf{Modality}: distribution of resource modalities among accessed content. \textbf{URL Overlap w/ Golden}: precision, recall, and F1 of the system's visited URLs against the golden reference URLs.}

\label{tab:search-comparison}
\end{table*}

\subsection{Decomposing the Performance Gap: Search vs.\ Reasoning}
\label{sec:gold_decomposition}
\paragraph{Setup.} To isolate whether performance limitations stem from the search stage or the reasoning stage, we conduct experiments using Gemini-3.1-Pro that progressively provide gold evidence (Table~\ref{tab:gold_decomposition}).
Starting from \agenticsearch, \textbf{+ Gold Sources Injection} injects gold source URLs into every web search response alongside live search results, while the agent retains all tools and must identify the gold URLs among noisy results.
\textbf{+ Gold Sources Only} removes web search entirely and provides only gold URLs, with tools still available for processing.
\textbf{Gold Sources Prompting} bypasses the agent framework entirely: gold videos and images are provided as native multimodal inputs, and web pages are fetched via URL context, in a single forward pass with no tools.

\paragraph{Takeaways.} 
\agenticsearch \textbf{+ Gold Sources Injection} yields only a modest improvement (+3.3\%), suggesting that search \textit{availability} alone is insufficient---the agent must also correctly \textit{select} and \textit{prioritize} relevant sources among noisy web results.
\agenticsearch \textbf{+ Gold Sources Only} further improves accuracy (+2.1\%), confirming that real-world distractors actively degrade agent reasoning even when gold evidence is present.
\textbf{Gold Sources Prompting} yields an additional +2.2\%, revealing that even when gold sources are provided, the agent does not always call tools to deeply investigate them---instead relying on surface-level information from URL titles or snippets rather than thoroughly examining the source content. 
From open search to perfect gold evidence, the total accuracy gain is 7.6\% (40.1\% $\to$ 47.7\%), upper-bounding the cost of search-stage limitations.
However, even with perfect gold evidence, accuracy remains relatively low, indicating that while both search effectiveness and multimodal reasoning remain critical open challenges, improving reasoning capabilities is the more pressing bottleneck on \benchmark.

\begin{table}[t]
\centering
\small
\begin{tabular}{lccccc}
\toprule
\textbf{Setting} & \textbf{Web Search} & \textbf{Gold Sources} & \textbf{Agent Tools} & \textbf{Acc.} \\
\midrule
\nosearch & \xmark & \xmark & \xmark & $24.7_{\pm1.6}$ \\
\nativesearch & \cmark & \xmark & \xmark & $29.0_{\pm1.1}$ \\ \midrule
\agenticsearch & \cmark & \xmark & \cmark & $40.1_{\pm2.8}$ \\
\quad + \textit{Gold Sources Injection  } & \cmark & \cmark & \cmark & $43.4_{\pm3.8}$ \\
\quad + \textit{Gold Sources Only} & \xmark & \cmark & \cmark & $45.5_{\pm2.3}$ \\ \midrule
\textit{Gold Sources Prompting} & \xmark & \cmark & \xmark & $47.7_{\pm2.0}$ \\
\bottomrule
\end{tabular}
\caption{Isolating search vs.\ reasoning limitations for Gemini-3.1-Pro on \benchmark. \textbf{Web Search}: whether the agent can search the open web. \textbf{Gold Sources}: whether gold sources are provided. \textbf{Agent Tools}: whether the agent can use custom tools (\texttt{visit\_webpage}, \texttt{watch\_video}) to process evidence.}
\label{tab:gold_decomposition}
\end{table}

\subsection{Human Performance}
\label{analysis: human_eval}
We conduct a human evaluation to analyze performance and compare with agents in \benchmark.
We recruit five undergraduate students to answer a randomly selected subset of 50 \benchmark questions using standard web search, without AI assistance.
Annotators record their answer, total time spent, number of search queries, and every resource consulted along with its relevance, modality, and URL.
We analyze human behavior, error patterns, and the effect of time on performance.

\paragraph{Comparing Human and Agents' Search Behavior.}
As shown in \autoref{tab:search-comparison}, humans achieve 71.4\% accuracy, substantially outperforming both \agenticsearch (40.1\%) and \nativesearch (30.9\%) using Gemini-3.1-Pro.
Humans use far fewer resources, averaging 2.9 searches and 2.9 website visits, compared to 9.1 searches and 3.5 visits for \agenticsearch.
Humans also achieve substantially higher precision in visited URLs (38.1\% vs.\ 1.8\%), indicating more effective source selection.
Although the agentic system attains high recall (61.4\%) due to the sheer volume of URLs encountered, its low precision indicates that the vast majority of retrieved sources are irrelevant.
Moreover, humans rely on a balanced mix of modalities (53.2\% text, 28.2\% video, 18.5\% image), whereas model-based systems are heavily text-dominant (87.0\% text for the agentic system; 96.2\% for the native system), with minimal video or image use.

\paragraph{Effect of Time on Performance.}
A striking finding emerges when comparing accuracy under a five-minute budget (Acc@5min, where questions exceeding five minutes are counted as incorrect) to overall accuracy.
Humans benefit substantially from additional time: their Acc@5min is 59.2\%, rising to 71.4\% overall, a gain of 12.2 \%.
This indicates that humans can productively leverage extra time to solve harder questions that require deeper search.
In contrast, agents show minimal improvement from additional time.
The native system improves by only 1.3 \% (29.6\% to 30.9\%), and the agentic system by 6.1 \% (34.0\% to 40.1\%)---far less than the human gain despite comparable average completion times (4.0 min for the agentic system vs.\ 4.1 min for humans).
These results point to a fundamental limitation of current search-augmented agents: unlike humans, who efficiently identify high-quality sources and extract relevant information even on difficult, time-consuming questions, agents struggle to synthesize information effectively as they process more content over longer reasoning chains, gaining little from the additional computation.
This finding is consistent with the over-exploration pattern described in Section~\ref{sec:failure_modes}: rather than productively deepening their search on difficult questions as humans do, agents tend to issue redundant queries and process tangentially relevant content, failing to converge.

\paragraph{Human Error Analysis.}
To better understand the nature of human errors, we categorize each incorrect human response into one of four categories based on the type of error made (\cref{fig:human_error}).
Among the responses where annotators provided an incorrect answer, the errors are predominantly minor extraction mistakes.
\emph{Wrong Count} (43\%) captures cases where the annotator identified the correct source but miscounted by a small margin (e.g., off by one album cover or one second of video).
\emph{Right Source, Wrong Detail} (29\%) includes cases where the annotator found the correct resource but extracted the wrong detail, such as reading a value from the wrong moment in a video or answering a different aspect of a multi-hop question.
\emph{Partial/Imprecise Answer} (14\%) covers responses that were on the right track but insufficiently specific (e.g., ``conservation law'' instead of ``conservation of charge'').
Only 14\% of errors fall into \emph{Others}, representing genuinely incorrect answers.
These results indicate that humans are generally able to identify the correct source and reasoning path, but often fail to extract precise information---underscoring that the benchmark's difficulty lies in fine-grained multimodal information extraction rather than source discovery, and highlighting the need for search agents that can effectively assist with such tasks.

\section{Related Work}

\paragraph{Multimodal Search Benchmarks.}

Prior work has focused on developing multimodal, search-augmented evaluation benchmarks.
Many of these benchmarks either provide multimodal inputs or include explicit modality cues that guide search agents toward which modalities to retrieve~\citep{li2025mm, geng2026webwatcher, zhang2026browsecomp, jiang2025mmsearch, tao2026mmsearchplus}. This design limits the ability to assess whether search agents can independently identify and retrieve the appropriate modality (e.g., whether the model can select audio sources or transcripts when a question asks about `what someone says', even in the absence of explicit modality cues).
In addition, prior work often focuses on a limited subset of modalities, primarily text and images, while overlooking others such as video and audio, which are common in real-world queries~\citep{jia2025benchmarking, yan2025multimodal, tian2025crosscheckbenchdiagnosingcompositionalfailures}. This restricts the evaluation of search agents' ability to perform multimodal reasoning across diverse modalities. 
To address these limitations, we introduce \benchmark, which consists of natural language queries without explicit modality source cues and includes questions that require multi-hop reasoning over a broader range of modalities.

\paragraph{Benchmarks for Reasoning under Web Noise.}

Prior work in the text domain shows that ambiguous, conflicting, and incomplete multi-source information can significantly degrade model performance, highlighting the importance of handling web noise~\citep{wang2025retrievalaugmented, lee2024well, pan2023attackingopendomainquestionanswering}.
Similar challenges have been explored in multimodal settings, but most focus on scenarios where such complexity is synthetically introduced, or where conflicts are constructed over predefined evidence segments within curated multimodal corpora, limiting the diversity and realism of noise compared to a realistic open-web environment~\citep{tian2025crosscheckbenchdiagnosingcompositionalfailures, zhang2025robustmultimodallargelanguage, semnani2025detectingcorpuslevelknowledgeinconsistencies, yan2025multimodal, jia2025benchmarking, wu2025mitigating}.
There is also a line of work on benchmarks for search-augmented agents that operate in open-web settings~\citep{li2025mm, tao2026mmsearchplus, geng2026webwatcher, jiang2025mmsearch}, but do not explicitly analyze how web noise affects reasoning or how agents respond to it. They often focus on limited modalities, primarily text and images.
In contrast, \benchmark explicitly induces web noise and requires search agents to reason across diverse modalities, including video and audio.

\section{Conclusion}
We introduced \benchmark, a human-annotated benchmark for evaluating search-augmented agents on multimodal evidence retrieval and reasoning in noisy web environments. 
It uses natural language queries without modality cues, spans diverse modalities (including video and audio), and requires reasoning over noisy, conflicting, and incomplete web evidence.
Evaluating search agents powered by ten different LLMs across three search settings, we find that \benchmark is highly challenging: average accuracy is 22.3\%, with the best-performing configuration achieving only 40.1\%. 
Compared to humans, agents are both less accurate and less efficient; humans search fewer but more precise queries and leverage more diverse modalities.
These results highlight the importance of \benchmark as a benchmark for evaluating search agents in challenging and realistic settings.

\section*{Ethics Statement}

While our dataset is constructed from publicly available web content, which may contain private or sensitive information,  we mitigate these risks through human annotation and careful review. All data is screened to ensure that no private, biased, or harmful content is included.

\section*{Acknowledgments}
We would like to thank Nithin Sivakumaran, Tianyi Niu, Vu Hoang Thien An, Dylan Zhao, and Hanqi Xiao for their contributions to the human evaluation. This work was supported by  ONR Grant N00014-23-1-2356, ARO Award W911NF2110220, NSF-CAREER Award 1846185, NSF AI Engage Institute DRL2112635, Microsoft Agentic AI Research and Innovation (AARI) program, and a Google PhD Fellowship. The views contained in this article are those of the authors and not of the funding agency.

\bibliography{colm2026_conference}
\bibliographystyle{colm2026_conference}

\appendix
\section{Limitations}
\label{app:limitation}
Our benchmark relies on Google Search as the primary search engine, which may introduce biases specific to its ranking algorithms; future work could validate findings across multiple search engines. The dataset comprises 162 questions, which, while comparable to other expert-vetted diagnostic benchmarks, may not capture the full diversity of real-world multimodal queries. Additionally, web content is inherently dynamic---URLs may become unavailable or content may change over time, potentially affecting reproducibility. We plan to regularly update the dataset to address this.

\section{\benchmark{} Details} \label{app: benchmark_details}

\subsection{Data Collection Details}\label{app: data_collection_details}
\paragraph{Annotation Fields.}
For each question, annotators record:
(a) the ground-truth answer along with a detailed explanation of the reasoning steps;
(b) source URLs (e.g., webpages, videos, PDFs) used as supporting evidence for deriving the reference answer;
(c) the source type of each resource (text, image, or video\footnote{Here, video sources incorporate visual and audio modalities.});
(d) the multimodal role, indicating whether non-text evidence serves as the answer source or as a reasoning component;
(e) the reasoning type, indicating whether the question is multi-hop and whether it introduces multimodal conflict; and
(f) the source of the question, labeled as \textit{from scratch} if both the question and its supporting evidence are newly constructed, or by the name of the originating dataset (e.g., \textit{SealQA}) if the question or evidence is adapted from an existing source.

\paragraph{Question Design Examples.}
To illustrate the no-modality-cues requirement, consider the following: instead of \emph{``In the attached chart, what year did Safari surpass IE in global browser market share?''}, we ask \emph{``According to StatCounter data, in which year did Safari surpass IE in global browser market share?''}
This phrasing avoids referencing any specific visual or auditory content while still requiring non-text evidence (a chart) to answer correctly.

For cases adapted from existing datasets, we use their question--answer pair as one piece of evidence (i.e., one hop) and augment it with additional evidence to construct \textit{new} multi-hop questions.
To ensure broad coverage, annotators are encouraged to include non-text evidence from at least two different source types where possible.

\subsection{Quality Control Details} \label{app: quality_control_details}

We employ a rigorous multi-round human review process. After initial question construction, each question is reviewed by a second annotator to assess: (1) answer correctness, (2) question clarity, (3) question difficulty, and (4) non-text modality requirements. Questions that fail any stage are revised and re-validated through subsequent review rounds.

\paragraph{Answer Correctness.}
Annotators independently verify the ground-truth answer by re-deriving it from the cited sources, flagging any discrepancies.

\paragraph{Question Clarity.}
Annotators check that each question is unambiguous, self-contained, and free of grammatical errors.

\begin{table}[t]
\centering
    \begin{tabular}{l|ccc}
    \toprule
    \textbf{Source} & \textbf{\# Q} & \textbf{Avg Q Len} & \textbf{Avg \# Res} \\
    \midrule
    Existing Benchmarks & 42 & 23.4 & 2.1 \\
    From Scratch & 120 & 25.2 & 1.9 \\
    \midrule
    \textbf{Total} & 162 & 24.7 & 2.0 \\
    \bottomrule
    \end{tabular}
\caption{Dataset statistics for \benchmark{}. \textbf{\# Q}: number of questions; \textbf{Avg Q Len}: average question length in words; \textbf{Avg \# Res}: average number of gold resources per question.}
\label{tab:dataset-stats}
\end{table}

\begin{figure}[t]
    \centering
    \begin{subfigure}[b]{0.51\textwidth}
        \includegraphics[width=\textwidth]{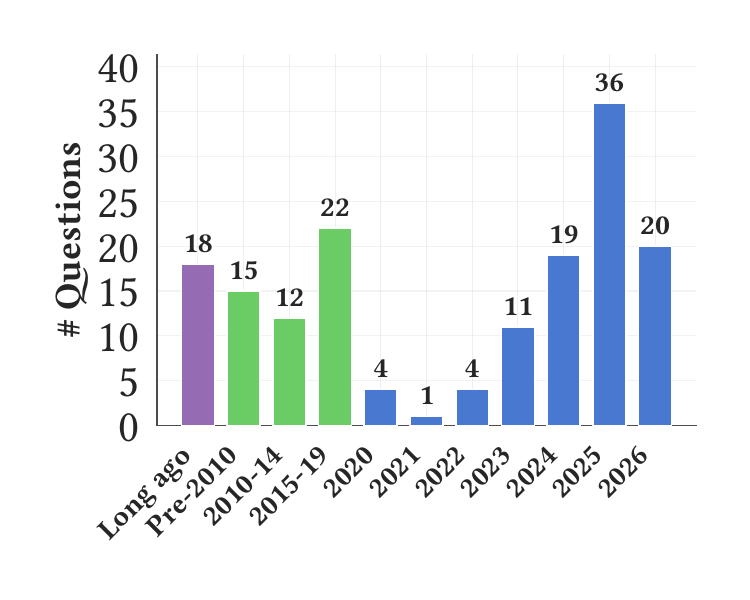}
        \caption{\textbf{Effective Year}}
        \label{fig:effective_year}
    \end{subfigure}
    \hfill
    \begin{subfigure}[b]{0.47\textwidth}
        \includegraphics[width=\textwidth]{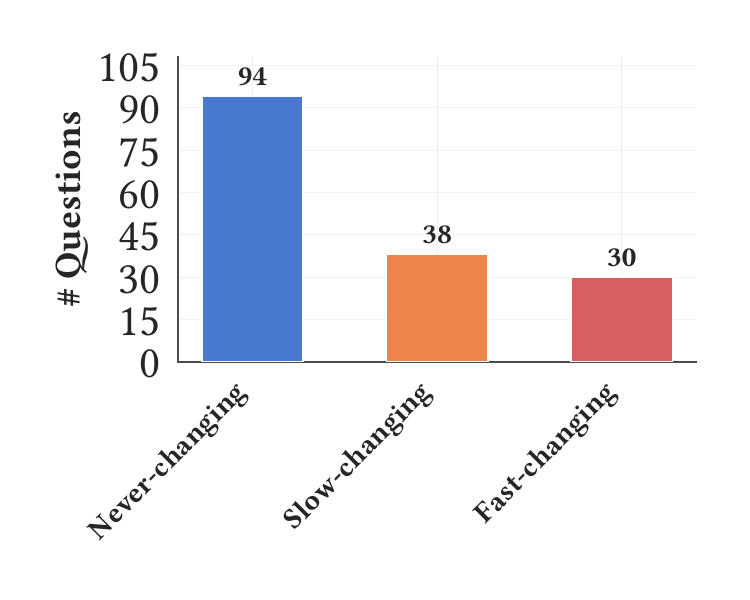}
        \caption{\textbf{Freshness}}
        \label{fig:freshness}
    \end{subfigure}
    \caption{Temporal characteristics of \benchmark{}. (a)~Distribution by \textbf{Effective Year} — the year in which the answer first became correct. Sparse years before 2020 are collapsed into Pre-2010, 2010--14, and 2015--19 buckets, while recent years are shown individually. (b)~Distribution by \textbf{Freshness} — how time-sensitive the ground-truth answer is.}
    \label{fig:temporal}
\end{figure}

\paragraph{Question Difficulty.}
We evaluate each question using ChatGPT's web interface~\citep{chatgptwebinterface} with web browsing enabled. Questions that are consistently answered correctly are revised or removed to maintain the desired level of challenge.

\paragraph{Non-Text Modality Verification.}
To verify that each question requires at least one non-text modality and cannot be solved with text alone, we apply a two-pass verification protocol:

\begin{enumerate}[noitemsep,topsep=0pt,leftmargin=*]
\item \textbf{Standard search pass:} The annotator decomposes each multi-hop question into constituent sub-questions. For each sub-question, the annotator attempts to answer it using text-only search via Google Search, simulating a standard retrieval setting.
\item \textbf{Adversarial search pass:} Given the ground-truth answer, the annotator queries the sub-question together with the answer string (e.g., submitting both \emph{``Who directed X?''} and the correct director name) to check whether any text-only document contains or implies the correct answer. This step is designed to uncover potential text-only shortcuts that a sufficiently capable retrieval system could exploit. We encourage the annotator to check as thoroughly as possible but limit it to up to 20 web searches.
\end{enumerate}

A question passes this check only if at least one sub-question cannot be resolved via text-only evidence under \textit{both} the standard and adversarial search passes.

\paragraph{Rejection Statistics.}
In the first round, approximately 39.5\% of initial candidates were rejected. Of the rejected questions, 45.3\% were successfully revised and accepted in the second round.

\begin{figure*}[ht]
\begin{tcolorbox}[width=0.99\textwidth, halign title=center, title = {Human Annotation Instruction}]
\small

\textbf{Goal.} Create challenging, multi-hop questions that require non-text evidence (images, videos, audio) to answer, designed to evaluate search-augmented agents.

\vspace{0.5em}
\textbf{Core Requirements.} Every question must satisfy all three:
\begin{enumerate}[noitemsep, topsep=2pt, leftmargin=*]
    \item \textbf{No modality cues.} Questions must be phrased in natural language without explicitly specifying the exact modality source (e.g., ``In the first episode of Rick and Morty Season 8, ...'', ``In this image...''). The question should read like a realistic user query.
    \item \textbf{Non-text evidence required.} At least one reasoning step must require non-text evidence (image, video, audio, chart, etc.) that cannot be resolved through text-only web search. This is verified through a two-pass protocol (see Quality Control).
    \item \textbf{Single unambiguous answer.} Each question must have exactly one correct, short, and verifiable answer.
\end{enumerate}

\vspace{0.5em}
\textbf{Classification Axes.} Annotate each question along two axes:
\begin{itemize}[noitemsep, topsep=2pt, leftmargin=*]
    \item \textit{Reasoning type} (one or both): \textbf{multi-hop} (combining information across multiple sources or modalities) and/or \textbf{multimodal conflict} (the question naturally triggers conflicting evidence across modalities in real search engines; do \textit{not} synthetically introduce conflicts).
    \item \textit{Multimodal role} (one or both): non-text evidence serves \textbf{as answer} (the answer can only be extracted from a non-text source) and/or \textbf{as reasoning chain} (non-text evidence provides an intermediate fact needed to derive the final answer).
\end{itemize}

\vspace{0.5em}
\textbf{Annotation Fields.} For each question, record:
\begin{itemize}[noitemsep, topsep=2pt, leftmargin=*]
    \item Ground-truth answer with a detailed explanation of the reasoning steps.
    \item Source URLs (webpages, videos, PDFs, images) used as supporting evidence.
    \item Source type of each resource (text, image, video/audio).
    \item Multimodal role and reasoning type labels (as defined above).
    \item Question origin: ``from scratch'' or the name of the originating dataset if adapted.
\end{itemize}

\vspace{0.5em}
\textbf{Tips for Question Construction.}
\begin{itemize}[noitemsep, topsep=2pt, leftmargin=*]
    \item Combine a text-retrievable fact (e.g., identifying an entity) with a visual detail that can only be found in an image or video (e.g., a color, count, or spatial arrangement).
    \item Avoid questions that can be answered by reading image captions, alt-text, or text surrounding the image on a webpage.
    \item Include diverse source types (text, images, video, audio) and diverse topics.
    \item Ensure the question is challenging: test it with ChatGPT with web browsing. If it is consistently answered correctly, revise or replace.
\end{itemize}

\end{tcolorbox}
\caption{Instruction for Human Annotation}
\label{fig:human_annotation}
\end{figure*}

\subsection{Human Annotation}
\label{app:human_annotation}
\autoref{fig:human_annotation} shows the human annotation guidelines.

\subsection{Data Statistics}
\cref{tab:dataset-stats} shows the overall data statistics. Beyond question and resource counts, we also annotate each question with two temporal dimensions: an Effective Year (the year in which the ground-truth answer first became valid) and a Freshness label (never-, slow-, or fast-changing), which together characterize how time-sensitive \benchmark{} is. \cref{fig:effective_year} shows the effective-year distribution: the bulk of the benchmark is concentrated in 2023--2026, with a long tail of older and time-agnostic (``Long ago'') questions. \cref{fig:freshness} shows the freshness distribution, which is dominated by never-changing questions (stable facts) but still includes a substantial fraction of slow- and fast-changing questions whose answers drift over time. 

\section{Experiments}
\subsection{Setup}
\label{app: exp: setup}

\paragraph{Search Setting Details.}
All closed-source models are evaluated via their official APIs. Supporting modalities and maximum context lengths for each model are listed in \cref{tab:modality_support}.

In the \textit{No Search} setting, models are evaluated without access to any tools.
In the \textit{Native Search} setting, we enable each model's built-in search capabilities. Recent LLM APIs are agentic by default, allowing models to autonomously invoke built-in tools and perform multi-turn reasoning within a single API call. For GPT models, we enable the \texttt{web\_search} tool, which supports searching the web, opening specific pages, and searching within pages, providing both retrieval and in-depth webpage understanding in a unified tool. For Gemini models, we enable both \texttt{Google Search} (web retrieval) and \texttt{URL Context} (webpage comprehension) to match the combined functionality of GPT's \texttt{web\_search} tool.

In the \textit{Agentic Multimodal Search} setting, we use a multimodal search agent framework built on smolagents~\citep{smolagents} that equips models with tools extending their effective modality coverage. In addition to the built-in \texttt{web\_search} tool (leveraging the Serper API for Google search), we incorporate two custom tools: \texttt{visit\_webpage}, which enhances the default webpage tool---limited to converting pages into markdown strings---by using Gemini-3-Flash with \texttt{URL Context} to interpret full webpage content including text and images; and \texttt{watch\_video}, which uses Gemini-3-Flash to directly process YouTube videos, enabling the agent to understand visual and audio content.

\begin{figure*}[ht]
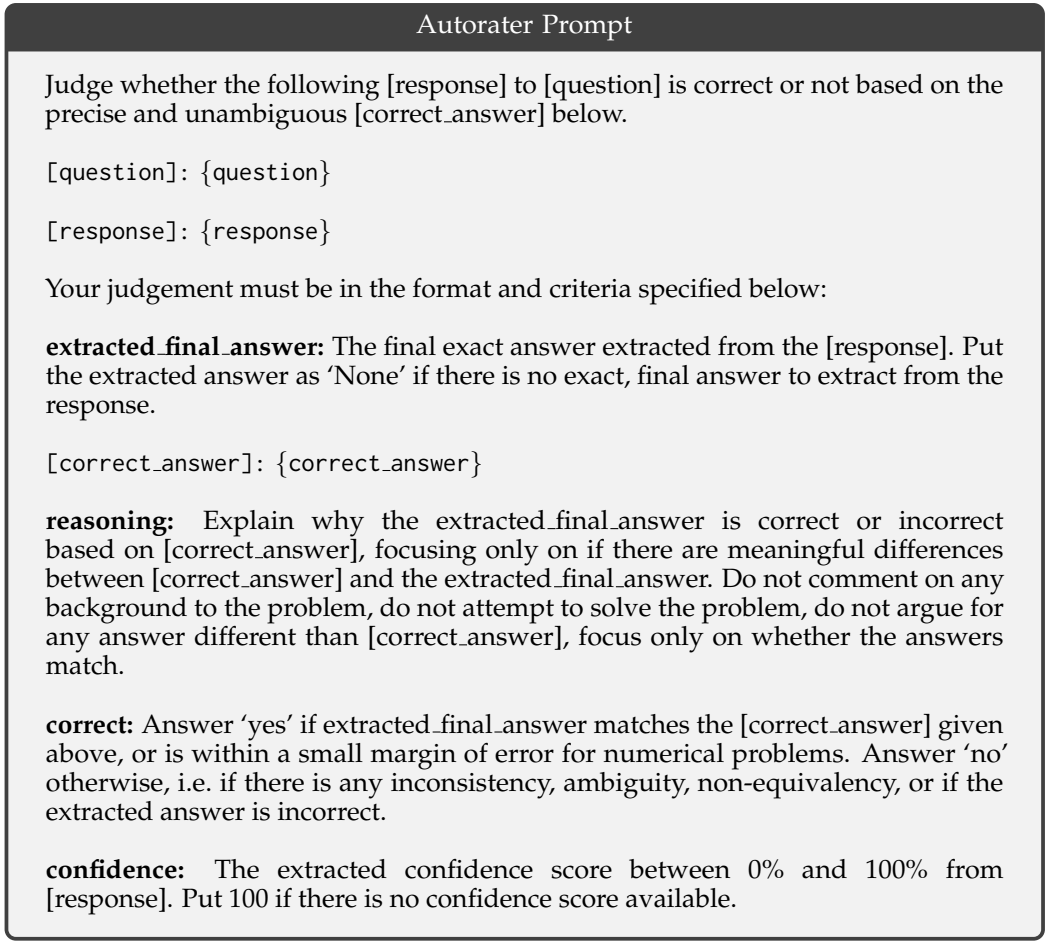

\begin{tcolorbox}[width=0.99\textwidth, halign title=center, title = {Autorater Prompt}
]
Judge whether the following [response] to [question] is correct or not based on the precise and unambiguous [correct\_answer] below. \newline \newline
\texttt{[question]:} \texttt{\{question\}} \newline \newline
\texttt{[response]:} \texttt{\{response\}} \newline \newline
Your judgement must be in the format and criteria specified below: \newline \newline
\textbf{extracted\_final\_answer:} The final exact answer extracted from the [response]. Put the extracted answer as `None' if there is no exact, final answer to extract from the response. \newline \newline
\texttt{[correct\_answer]:} \texttt{\{correct\_answer\}} \newline \newline
\textbf{reasoning:} Explain why the extracted\_final\_answer is correct or incorrect based on [correct\_answer], focusing only on if there are meaningful differences between [correct\_answer] and the extracted\_final\_answer. Do not comment on any background to the problem, do not attempt to solve the problem, do not argue for any answer different than [correct\_answer], focus only on whether the answers match. \newline \newline
\textbf{correct:} Answer `yes' if extracted\_final\_answer matches the [correct\_answer] given above, or is within a small margin of error for numerical problems. Answer `no' otherwise, i.e.\ if there is any inconsistency, ambiguity, non-equivalency, or if the extracted answer is incorrect. \newline \newline
\textbf{confidence:} The extracted confidence score between 0\% and 100\% from [response]. Put 100 if there is no confidence score available.
\end{tcolorbox}
\caption{Autorater prompt used for grading responses. Placeholders \texttt{\{question\}}, \texttt{\{response\}}, and \texttt{\{correct\_answer\}} are filled at evaluation time.}
\label{fig:acc_prompt}
\end{figure*}

\paragraph{Evaluation Details.}
We evaluate using an LLM-as-judge with the same prompt as BrowseComp~\citep{wei2025browsecomp}. Prompt can be seen in \autoref{fig:acc_prompt}.

\begin{table}[t]
\small
\centering
\begin{tabular}{l|c|cccc|cccc}
\toprule
&  & \multicolumn{4}{c|}{Input Query} & \multicolumn{4}{c}{Built-in Search} \\ 
\midrule
\textbf{Model} & Context (In/Out) & Text & Image & Video & Audio & Text & Image & Video & Audio\\
\midrule
GPT & 400k/128k & \cmark & \cmark & \xmark & \xmark& \cmark & \cmark & \xmark & \xmark \\
Gemini & 1M/64k & \cmark&\cmark&\cmark&\cmark&\cmark&\cmark& \xmark & \xmark \\
\midrule
Qwen3 & 230k/32k & \cmark&\xmark&\xmark&\xmark & - & - & - &-\\
\bottomrule
\end{tabular}
\caption{
Model context window sizes and modality support. \textbf{Context (In/Out)}: maximum input/output token limits. \textbf{Input Query}: modalities the model can accept as direct input. \textbf{Built-in Search}: modalities the model can process when using its built-in search tool under \nativesearch.
}
\label{tab:modality_support}
\end{table}

\begin{figure*}[ht]
\begin{tcolorbox}[width=0.99\textwidth, halign title=center, title = {Human Evaluation Instruction}]

\textbf{Overview} \\
You will be presented with a set of questions in a Google Sheets spreadsheet and asked to find the correct answer to each using the open web. Each question has a \textbf{single, unambiguous short answer} (e.g., a name, number, date, or brief phrase).

\vspace{6pt}
\textbf{How to Use the Spreadsheet} \\
Each row in your assigned Google Sheet contains one question. For each row, fill in the following columns:

\vspace{4pt}
\begin{tabular}{@{}p{0.22\textwidth} p{0.72\textwidth}@{}}
\toprule
\textbf{Column} & \textbf{What to enter} \\
\midrule
Question & Already filled in — read carefully. \\
Your Answer & Your short answer to the question. \\
Annotation Time & How long it took you to find the answer (e.g., ``2 min'', ``4 min''). \\
Number of Queries & The total number of search queries you made for this question. \\
Resource 1, 2, \ldots & Every resource (webpage, image, video, etc.) you opened while searching. For each, record three things separated by commas: (1) whether the resource was \textbf{relevant} or \textbf{not relevant}, (2) the modality (e.g., \texttt{text}, \texttt{image}, \texttt{video}, \texttt{table}), and (3) the URL. \\
\bottomrule
\end{tabular}

\vspace{6pt}
\textbf{Your Task}
\begin{enumerate}[leftmargin=*, nosep]
    \item \textbf{Read the question carefully.} Each question is written in plain natural language.
    \item \textbf{Search the web to find the answer.} Use Google Search for entering queries and find websites, videos, etc.\ to find relevant information.
    \item \textbf{Do NOT close any tabs while searching.} Keep all tabs open throughout your search so that you can accurately record all resources and queries at the end.
    \item \textbf{Record your answer} in the ``Your Answer'' column.
    \item \textbf{Record the time} in minutes it took you in the ``Annotation Time'' column.
    \item \textbf{Count your search queries.} After finding the answer (or giving up), go through your browser history/tabs and count the total number of search queries you made.
    \item \textbf{Record ALL resources.} Go through every tab you opened and record each one in the Resource columns — not just the ones that contained the answer. For each resource, indicate whether it was \textbf{relevant} or \textbf{not relevant}, the modality, and the URL.
\end{enumerate}

\vspace{6pt}
\textbf{Answer Format}
\begin{itemize}[leftmargin=*, nosep]
    \item Keep your answer \textbf{short and precise} — typically a single entity, value, or brief phrase.
    \item Do \textbf{not} include full sentences or explanations unless specifically requested.
\end{itemize}

\vspace{6pt}
\textbf{Important Notes}
\begin{itemize}[leftmargin=*, nosep]
    \item \textbf{Do NOT use any AI tools.} Do not use ChatGPT, Google Search AI mode (AI Overviews), Bing Copilot, Perplexity, or any other AI-powered assistant.
    \item \textbf{Do NOT close your tabs.} Keep all tabs open while working on each question.
    \item Examine all resources carefully and \textbf{record every resource}, even unhelpful ones.
    \item \textbf{Verify before submitting.} Cross-check your answer with at least one additional source when possible.
    \item If you cannot find the answer, write ``Unanswerable'' and briefly note what you tried.
\end{itemize}

\end{tcolorbox}
\caption{Instruction for Human Evaluation}
\label{fig:human_eval}
\end{figure*}

\section{Human Evaluation}\label{sec: human_eval}
Human evaluation guidelines can be found in \autoref{fig:human_eval}.

\begin{table}[t]
  \centering
  \resizebox{\textwidth}{!}{
  \begin{tabular}{lcccc}
  \toprule
  \textbf{Model} & \textit{No Search} & \textit{Native Search} & \textit{Native Search + Video Tool} & \textit{Agentic Multimodal Search} \\
  \midrule
  Gemini 3 Flash & $19.1_{\pm3.2}$ & $31.7_{\pm3.8}$ & $36.8_{\pm2.6}$ & $32.9_{\pm0.9}$ \\
  Gemini 3 Pro & $23.5_{\pm1.1}$ & $28.8_{\pm1.4}$ & $35.6_{\pm0.4}$ & $39.9_{\pm1.6}$ \\
  Gemini 3.1 Lite & $12.8_{\pm2.3}$ & $20.6_{\pm2.2}$ & $21.6_{\pm1.2}$ & $26.3_{\pm1.9}$ \\
  Gemini 3.1 Pro & $24.7_{\pm1.6}$ & $29.0_{\pm1.1}$ & $37.5_{\pm2.0}$ & $40.1_{\pm2.8}$ \\
  \bottomrule
  \end{tabular}}
  \caption{Impact of adding a video processing tool to \nativesearch. Accuracy (\%) across four Gemini models under four settings: \textit{No Search}, \textit{Native Search}, \textit{Native Search + Video Tool} (adding video processing tool), and \agenticsearch (full multimodal agent).
  }
  \label{tab:gemini_video_tool}
\end{table}

\section{Analysis}

\subsection{Impact of Adding Video Processing Tool}
\label{app: ablation: video_process}

As shown in \cref{tab:gemini_video_tool}, adding a video processing tool to \nativesearch consistently improves accuracy, with gains ranging from +1.0\% (Gemini 3.1 Lite) to +8.5\% (Gemini 3.1 Pro), averaging +5.7\% across agents.
This confirms that video evidence is critical for a substantial portion of \benchmark questions and that \nativesearch's inability to process video is a significant limitation.
Comparing \nativesearch with the video tool to \agenticsearch, we observe that \agenticsearch still outperforms on three of four agents (Gemini 3 Pro: 39.9\% vs.\ 35.6\%, Gemini 3.1 Lite: 26.3\% vs.\ 21.6\%, Gemini 3.1 Pro: 40.1\% vs.\ 37.5\%), except for Gemini 3 Flash (32.9\% vs.\ 36.8\%).
We attribute this gap to two factors: (1) \nativesearch sometimes fails to invoke the video tool, whereas \agenticsearch proactively calls \texttt{watch\_video}; and (2) \agenticsearch locates more relevant videos through its dedicated \texttt{search\_video} tool, while \nativesearch relies on the built-in Google Search, which sometimes retrieves irrelevant videos.

\end{document}